\documentclass{article}

\usepackage[left=3cm, right=3cm, top=2cm, bottom = 2cm]{geometry}

\usepackage{scalerel}
\usepackage{tikz}
\usetikzlibrary{svg.path}

\definecolor{orcidlogocol}{HTML}{A6CE39}
\tikzset{
  orcidlogo/.pic={
    \fill[orcidlogocol] svg{M256,128c0,70.7-57.3,128-128,128C57.3,256,0,198.7,0,128C0,57.3,57.3,0,128,0C198.7,0,256,57.3,256,128z};
    \fill[white] svg{M86.3,186.2H70.9V79.1h15.4v48.4V186.2z}
                 svg{M108.9,79.1h41.6c39.6,0,57,28.3,57,53.6c0,27.5-21.5,53.6-56.8,53.6h-41.8V79.1z M124.3,172.4h24.5c34.9,0,42.9-26.5,42.9-39.7c0-21.5-13.7-39.7-43.7-39.7h-23.7V172.4z}
                 svg{M88.7,56.8c0,5.5-4.5,10.1-10.1,10.1c-5.6,0-10.1-4.6-10.1-10.1c0-5.6,4.5-10.1,10.1-10.1C84.2,46.7,88.7,51.3,88.7,56.8z};
  }
}

\newcommand\orcidicon[1]{\href{https://orcid.org/#1}{\mbox{\scalerel*{
\begin{tikzpicture}[yscale=-1,transform shape]
\pic{orcidlogo};
\end{tikzpicture}
}{|}}}}

\usepackage[utf8]{inputenc}
\usepackage[T1]{fontenc}
\usepackage{lmodern}
\usepackage[UKenglish]{isodate}

\usepackage{amsmath}
\usepackage{graphicx}
\usepackage{xcolor}
\usepackage{tikz}
\usepackage{textcomp}

\usepackage{cite}

\usepackage[hidelinks]{hyperref}
\usepackage[strings]{underscore}

\usepackage{tabularx}
\usepackage{paralist}

\usepackage{zed-csp}

\usepackage[xindy]{glossaries}

\newacronym{rml}{RML}{Runtime Monitoring Language}

\newacronym{ltl}{LTL}{Linear-time Temporal Logic}

\newacronym{ctl}{CTL}{Computation Tree Logic}

\newacronym{ptl}{PTL}{Probabilistic Temporal Logic}

\newacronym{pctl}{PCTL}{Probabilistic Computation Tree Logic}

\newacronym{pctl*}{PCTL*}{Probabilistic Computation Tree Logic*}

\newacronym{tctl}{TCTL}{Timed Computation Tree Logic}

\newacronym{tltl}{TLTL}{Timed Linear-time Temporal Logic}

\newacronym{rvltl}{RV-LTL}{Runtime Verification Linear Temporal Logic}

\newacronym{stl}{STL}{Signal Temporal Logic}

\newacronym{ltl-x}{LTL-X}{a version of Linear Time Logic without the `next' operator}

\newacronym{mtl}{MTL}{Metric Temporal Logic}

\newacronym{iactlk-x}{IACTLK-X}{a fragment of the temporal-epistemic logic CTLK, without the `next' operator}

\newacronym{fotl}{FOTL}{First-Order Temporal Logic}

\newacronym{bdi}{BDI}{Belief-Desire-Intention}

\newacronym{prs}{PRS}{Procedural Reasoning System}

\newacronym{are}{ARE}{Autonomous Reasoning Engine}

\newacronym{dmars}{dMARS}{distributed Multi-Agent Reasoning System}

\newacronym{fsm}{FSM}{Finite-State Machine}

\newacronym{pfsm}{PFSM}{Probabilistic Finite-State Machine}

\newacronym{apfsm}{APFSM}{\textit{Autonomous} Probabilistic Finite-State Machine}

\newacronym{pha}{PHA}{Probabilistic Hybrid Automata}

\newacronym{fsa}{FSA}{Finite-State Automata}

\newacronym{ta}{TA}{Timed Automata}

\newacronym{gspn}{GSPN}{Generalised Stochastic Petri Net}

\newacronym{mopn}{MOPN}{Marked Ordinary Petri Net}

\newacronym{ba}{BA}{B\"{u}chi Automata}

\newacronym{mdp}{MDP}{Markov Decision Process}

\newacronym{dtmc}{DTMC}{Discrete-Time Markov Chain}

\newacronym{pars}{PARS}{Process Algebra for Robot Schemas}

\newacronym{fsp}{FSP}{Finite State Processes}

\newacronym{piadl}{$\pi$ADL}{$\pi$ calculus combined with ADL}

\newacronym{csp}{CSP}{Communicating Sequential Processes}

\newacronym{ccs}{CCS}{Calculus of Communicating Systems}

\newacronym{wsccs}{WSCCS}{Weighted Synchronous CCS}

\newacronym{csp|b}{CSP$\|$B}{an integrated formalism combining CSP and B}

\newacronym{fdr}{FDR}{the Failures-Divergences Refinement checker}

\newacronym{jpf}{JPF}{Java PathFinder}

\newacronym{ajpf}{AJPF}{Agent Java PathFinder}

\newacronym{mcmas}{MCMAS}{Model Checker for Multi-Agent Systems}

\newacronym{ltlmop}{LTLMoP}{Linear Temporal Logic MissiOn Planning}

\newacronym{adl}{ADL}{Architecture Description Language}

\newacronym{aadl}{AADL}{Architecture Analysis and Design Language}

\newacronym{lts}{LTS}{Labelled-Transition System}

\newacronym{uml}{UML}{Unified Modelling Language}

\newacronym{sysml}{SySML}{Systems Modelling Language}

\newacronym{robotml}{RobotML}{Robot Modelling Language}

\newacronym{dsl}{DSL}{Domain Specific Language}

\newacronym{sct}{SCT}{Supervisory Control Theory}

\newacronym{ide}{IDE}{Integrated Development Environment}

\newacronym{mde}{MDE}{Model-Driven Engineering}

\newacronym{ai}{AI}{Artificial Intelligence}

\newacronym{fdir}{FDIR}{Fault Detection, Isolation and Recovery}

\newacronym{ifm}{iFM}{Integrated Formal Methods}

\newacronym{ros}{ROS}{Robot Operating System}

\newacronym{dl}{\text{\upshape\textsf{d{\kern-0.05em}L}}}{differential dynamic logic}

\newacronym{vm}{VM}{Virtual Machine}

\newacronym{bip}{BIP}{Behaviour Interaction Priority}

\newacronym{fol}{FOL}{First-Order Logic}

\newacronym{owl}{OWL}{Web Ontology Language}

\newacronym{ria}{RIA}{Restore Invariant Approach}

\newacronym{ocl}{OCL}{Object Constraint Language}

\newacronym{sil}{SIL}{Safety Integrity Level}

\newacronym{rule}{RULE}{Robotic Urban-Like Environment}

\newacronym{forms}{FORMS}{FOrmal Reference Model for Self-adaptation}

\newacronym{rv}{RV}{Runtime Verification}

\newacronym{prv}{PRV}{Predictive Runtime Verification}

\newacronym{mmprv}{MMPRV}{Multi-Model Predictive Runtime Verification}

\newacronym{cmmprv}{CMMPRV}{Centralised Multi-Model Predictive Runtime Verification}

\newacronym{dmmprv}{DMMPRV}{Decentralised Multi-Model Predictive Runtime Verification}

\newacronym{sua}{SUA}{System Under Analysis}

\newacronym{race}{RACE}{Remote Applications in Challenging Environments}

\newacronym{ccfe}{CCFE}{Culham Centre for Fusion Energy}

\newacronym{ukaea}{UKAEA}{UK Atomic Energy Authority}

\newacronym{jet}{JET}{Joint European Torus}

\newacronym{homm}{$HOMM$}{$HANDS\_ON\_MODE\_MONITOR$}
\newacronym{ssm}{$SSM$}{$SAFE\_SPEED\_MONITOR$} 

\newacronym{homm-text}{HOMM}{Hands-On Mode Monitor}
\newacronym{ssm-text}{SSM}{Safe Speed Monitor}

\usepackage[disable]{todonotes}

\usepackage{lineno}

\usepackage{pdfpages}

\newcommand{\Varanus}{\textsc{Varanus}}

\makeglossaries

\begin{document}
\title{Offline Runtime Verification of Safety Requirements using CSP\thanks{Work supported through the UKRI RAIN grant EP/R026084, and enabled by previous work carried out within the framework of the EUROfusion Consortium, which has received funding from the Euratom research and training programme 2014‐2018 and 2019‐2020, grant No 633053. The views and opinions expressed herein do not necessarily reflect those of the European Commission. Thanks go to Luigi Pangione and Rob Skilton at CCFE; Matt Webster for helpful discussion; and Angelo Ferrando, Louise Dennis, and Marie Farrell for comments on early drafts. Most of this work was done when the author was employed by the University of Liverpool, UK}.}

\author{Matt Luckcuck\orcidicon{0000-0002-6444-9312} \\ \small{Department of Computer Science, Maynooth University, Ireland  } }
\date{\today}

\maketitle 

\begin{abstract}

Dynamic formal verification is a key tool for providing ongoing confidence that a system is meeting its requirements while in use, especially when paired with static formal verification before the system is in use. 
This paper presents a workflow and \gls{rv} toolchain, \Varanus, and their application to an industrial case study. Using the workflow we manually derive a \gls{csp} model from natural-language safety requirements documents, which {\Varanus} uses as the monitor oracle. This reuse of the model means that the monitor oracle does not have to be developed separately, risking inconsistencies between it and the model for static verification. The approach is demonstrated by the offline \gls{rv} of a teleoperated manipulation system, called MASCOT, which enables remote operations inside the \gls{jet} fusion reactor. We describe our model of the MASCOT safety design documents (including how the modelling process revealed an underspecification in the design) and evaluate the {\Varanus} toolchain's utility. 
The workflow and tool provide validation of the safety documents, traceability of the safety properties from the documentation to the system, and a verified oracle for \gls{rv}.

\end{abstract}

\glsresetall


\section{Introduction}
\label{sec:intro}



\gls{rv} provides ongoing confidence that a system continues it meet its requirements after it has been developed and is in use. These requirements are often expressed in natural-language safety documents, based on trusted (often non-formal) safety analysis techniques. Integrating formal methods with existing non-formal safety techniques is useful, often necessary~\cite{farrell2018}, and provides another tool for the verification toolbox. 

This paper presents an \gls{rv} workflow and toolchain, where the oracle (the component that provides the verdict on whether or not the \gls{sua} violates the specification) is a \gls{csp} model of the behaviour described in the system's design. In our workflow, we build this oracle (by hand) from existing natural-language safety documents, capturing both the behaviour and the safety requirements, then verify the model of the behaviour against the safety requirements using the \gls{csp} model checker, \gls{fdr}\cite{fdr}. We then use this model as the monitoring oracle, again using \gls{fdr}. Formally modelling and verifying the safety documents can expose errors, we discuss an instance of this in \S\ref{sec:discussion}.

Our \gls{rv} approach uses a \gls{csp} model directly, as the \gls{rv} oracle; whereas related approaches use implementations or dialects of \gls{csp} (see \S\ref{sec:relatedWork}). This means that the model has two uses in our workflow: first in model checking a system's design, to identify faults; and second as the \gls{rv} oracle, to verify that the system continues to obeys its safety requirements. Reusing the model supports traceability of the safety properties from the safety documentation to a system artefact, which has been formally verified so it also gives strong evidence that the correct safety properties are being monitored.

We illustrate our approach using offline \gls{rv} of the MASCOT system (described in \S\ref{sec:mascot}) which is a pair of  master-slave robotic arms that enable engineers to operate remotely inside a fusion reactor. For the upgrade to MASCOT version 6, which includes some autonomous movements, a new safety analysis and design were produced. These safety documents are the source material for our formal model and safety properties. The key concerns of the safety documents are the speed of the master arms (which could collide with the operator) and managing explicit changes between hands-on and autonomous modes of operation.

The rest of the paper is laid out as follows; \S\ref{sec:relatedWork} describes related work, and \S\ref{sec:approach} describes our toolchain and workflow. In \S\ref{sec:mascot} we describe the MASCOT system, its \gls{csp} model, and how we validated and verified the model. \S~\ref{sec:eval} describes how we evaluate the toolchain. In \S\ref{sec:discussion} we discuss the development of this approach and toolchain. Finally, \S\ref{sec:conclusion} summarises our approach and describes future work.

\section{Related Work}
\label{sec:relatedWork}

Our approach uses a \gls{csp} model directly as the \gls{rv} oracle. There are other approaches to \gls{rv} that make use of \gls{csp}-style notations, as dialects or implementations of \gls{csp}. For example, the \textbf{\textsf{Jass}} system\cite{bartetzko_jass_2001} provides an assertion language for Java programs (developed before Java introduced its \texttt{assert} statement). \textbf{\textsf{Jass}} includes an assertion for specifying the permissible traces of method invocations, using a \textit{dialect} of \gls{csp}~\cite{moller_specifying_2002}. The dialect is adapted to make it easier to specify the requirements of Java programs, which means that a \gls{csp} model would need to be manually translated into the dialect. While this does not seem onerous, it is another manual step to add to the workflow. Further, \textbf{\textsf{Jass}} is specialised to Java programs, whereas the work we present in this paper is agnostic of the implementation language of the \gls{sua}.

Another indirect use of the \gls{csp} language is $CSP_E$~\cite{yamagata_runtime_2016}, which is a shallow-embedded \gls{dsl} in Scala, specifically for \gls{rv}. Again, this would require another manual conversation to be added to the workflow. Further, $CSP_E$ is missing elements of \gls{csp} that would make the conversion from standard \gls{csp} less straightforward than reusing the model.
 

Dynamic verification approaches have been applied to other robotic examples where speed and potential collision with humans is of concern. For example an approach that calculates the reachable sets, where a collision is possible between a mobile robot and humans in its environment~\cite{Liu2017}. The calculation is based on kinematic models of the robot and humans, and checks that the robot cannot enter an area that a pedestrian could reach at the same time. This approach has also been applied to robot arms that share a workspace with a human~\cite{Pereira2015}. Both of these are online techniques, but do not also offer static validation. 

By contrast, ModelPlex\cite{Mitsch2016} combines offline (static) verification of a model of a cyber-physical system with online validation of the system's runtime behaviour against the model. If the system deviates from is verified model, then ModelPlex triggers safe fallback behaviour. This approach uses hybrid models, written in \gls{dl}, and automatically synthesises the runtime monitors. Another example of this combination of static verification and dynamic \gls{rv} checks an autonomous robot against its model of its deployment environment~\cite{ferrando_recognising_2018}. The approach highlights where interactions with the real world invalidate the design-time assumptions about the robot's environment. Both of these approaches are similar in intent to our work, but our case study is human-controlled and does not use a model of its environment.

\section{Toolchain and Workflow}
\label{sec:approach}

This section describes our \gls{rv} toolchain and its intended workflow. Our toolchain, {\Varanus}\footnote{{\Varanus}, named for the biological genus of Monitor Lizards, is available at \url{https://github.com/autonomy-and-verification/varanus/tree/FMICS-Data}.} uses \gls{fdr}'s API to check the behaviour of the \gls{sua} against a \gls{csp} model of its safety system. This paper presents an example of offline \gls{rv}, where {\Varanus} read events from a file; but it can also listen for events over a socket or WebSocket, for online \gls{rv}. {\Varanus} constructs a trace from the events, asks \gls{fdr} if it is a valid trace of the model, and returns the verdict (pass or fail). {\Varanus} produces a log of the check and the response time, used in \S\ref{sec:eval}.

\begin{figure}[t]
\centering
\includegraphics[width=\textwidth]{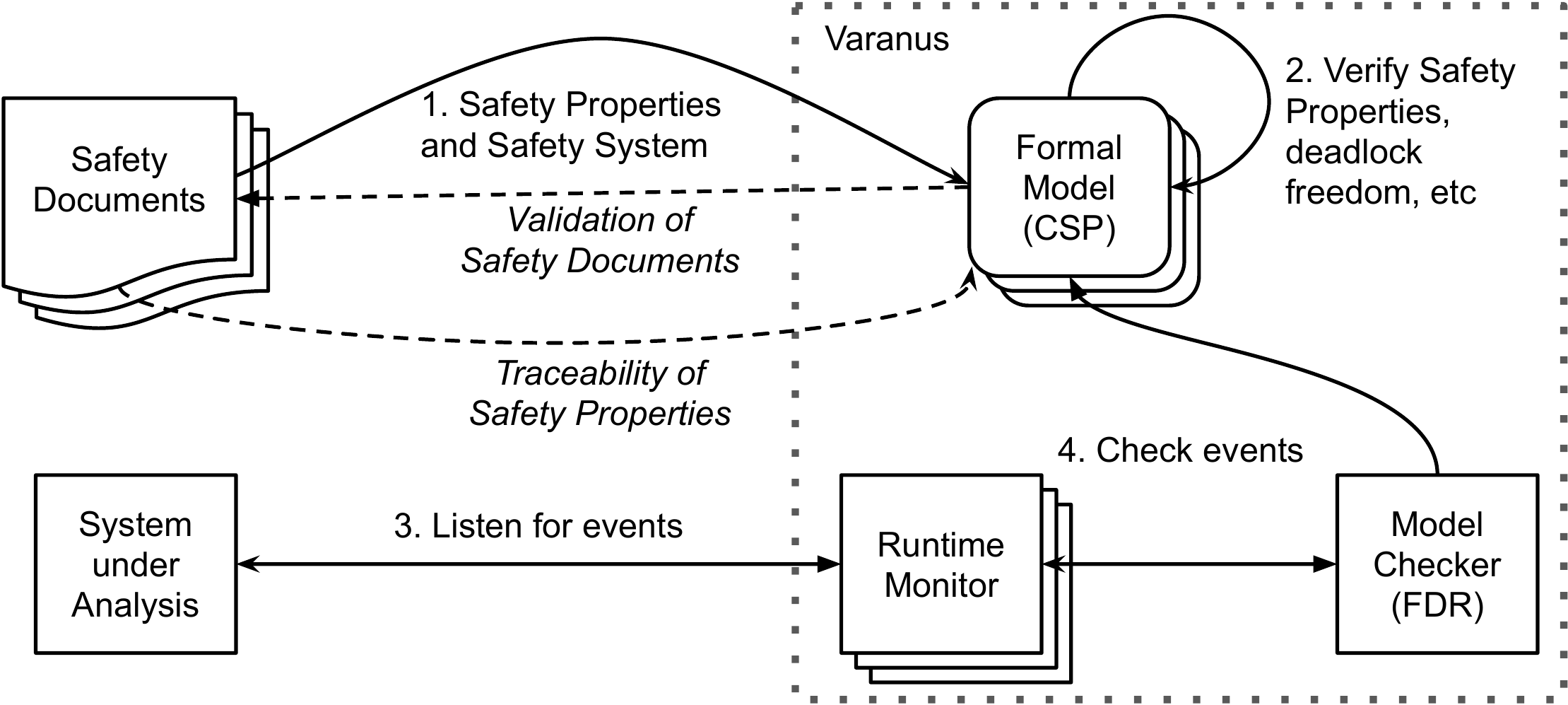}
\caption{Diagram of the {\Varanus} toolchain, with numerical order of the workflow. Dashed box is the {\Varanus} system boundary, rounded-edge boxes represent the model, and squared-edge boxes are programs. The arrowheads show direction of data flow. \label{fig:approach}} 
\end{figure}
Figure~\ref{fig:approach} illustrates the {\Varanus} toolchain and workflow. In step 1 we formalise the system's safety properties and functions, extracted from safety documents (see \S~\ref{sec:step1}). Step 2 verifies the model against the safety properties, validating the model (see \S\ref{sec:step2}). Then, in step 3, {\Varanus} listens for events emitted by the \gls{sua} (see \S\ref{sec:step3}). Finally, step 4 checks the events against the model (using \gls{fdr}) to determine if the system is performing the safety functions (see \S\ref{sec:step4}).



\subsection{Step 1: Formalising}
\label{sec:step1}

This step involves modelling the behaviour and the safety requirements in the system's safety documents, both the behaviour and the requirements become \gls{csp} processes. This step was performed manually, so required careful reading of the natural-language descriptions to formalise them. This can be time-consuming and requires formal modelling expertise. Discussion in the literature makes it clear that building specifications is still the biggest bottleneck in formal methods~\cite{rozier2016}. It is vital that the model built in this step is correct because it is the \gls{rv} oracle in the {\Varanus} toolchain.

Using \gls{csp} enables us to model both the behaviour and requirements described in the safety documents using events. For example, the MASCOT safety subsystem monitors the speed of the master arms and initiates a protective stop if a speed limit is exceeded, which  {we model with the $speed$ and $protective\_stop$ events. We describe \gls{csp} and our modelling approach in \S\ref{sec:modelling}. There are examples of requirements in the safety design and their \gls{csp} translation in \S\ref{sec:modelVerification}. In the author's experience, software engineering or safety specifications are often written in a way that enables smooth translation into \gls{csp}, as discussed in \S~\ref{sec:discussion}.





The system's safety requirements, \textit{safety properties}, are formalised as \gls{csp} processes. The next step (\S\ref{sec:step2}) describes how we verify that the model preserves these safety properties. This direct link between the safety documents and our model supports traceability of the requirements into a formal artefact (our model) of the system's development.

\subsection{Step 2: Verifying}
\label{sec:step2}

This step takes the \gls{csp} model and safety properties from \S\ref{sec:step1} and verifies that the model preserves the properties. Since both the properties and model are written in \gls{csp}, this step uses the \gls{csp} model checker, \gls{fdr}. 

We use \gls{fdr} to debug the model, by a combination of built-in assertions (deadlock, determinism, and divergence), and its Probe tool (which allows a user to step through a process's available events). Then, the model can be verified against the safety properties. Both of these processes are iterative, and verification may highlight more areas to debug.

Verifying that the model preserves the safety properties has the side-benefit of validating the safety document itself. Formalising the natural-language specification of the safety system's behaviour can highlight inconsistencies in the document. Even simple verification checks, such as checking for deadlock-freedom, can provide counterexamples showing an error in the safety document. An instance of this occurred during this work, which we discuss in \S\ref{sec:modelVerification}. 

\subsection{Step 3: Listening}
\label{sec:step3}

In this step, {\Varanus} listens to and translates events of the \gls{sua} into events of the \gls{csp} model. As previously mentioned, {\Varanus} can read a file of recorded events, for offline \gls{rv}; or connect to a running system, for online \gls{rv}.
Each \gls{sua} will require a specific mapping between its events and those in the model; and an implementation of the \texttt{system_interface}, which connects to the \gls{sua} or file. Step 2 (\S\ref{sec:step2}) provides confidence that the model accurately represents the safety system (and therefore the \gls{sua}'s behaviour), so that it can be used for \gls{rv}. 


{\Varanus} stores a trace of the events generated by the \gls{sua}, translated for the \gls{csp} model. The trace begins with the model's initial event,and for each new event {\Varanus} converts the \gls{sua} event into a \gls{csp} event, using the user-supplied mapping, and appends it to the trace. 
 In the offline case, {\Varanus} converts the whole trace before checking (described in \S\ref{sec:step4}). In the online case the trace is checked after each event is appended, which is a limitation of the current approach, because the \gls{fdr} API was not designed for \gls{rv}, and is why we focus on offline \gls{rv} in this paper.


\subsection{Step 4: Checking}
\label{sec:step4}

{\Varanus} uses \gls{fdr}'s built-in $[has~trace]$ assertion to check if the \gls{sua}'s trace (from \S\ref{sec:step3}) is a valid trace of the model; that is, that the system is behaving according to the model. Each check is built from the template: 

\centerline{$assert~MODEL:[has~trace]:\langle sua\_trace \rangle$}

\noindent where $MODEL$ is the \gls{csp} process that defines the safety model, and $sua\_trace$ is a trace of events from the \gls{sua}. 
\gls{fdr} checks that the $MODEL$ can perform the $sua\_trace$ without diverging or refusing its events. This means that if $MODEL$ offers a choice of events, each option will be explored.

\gls{fdr} returns that the assertion check has either passed or failed, and provides a counterexample for a failing result. In our case study (\S\ref{sec:mascot}), {\Varanus} passes  information is returned to the user. In online \gls{rv} examples, this information could be used by the \gls{sua} for replanning or recovery.



\section{Case Study: The MASCOT System}
\label{sec:mascot}

This section describes an application of the {\Varanus} toolchain (\S\ref{sec:approach}). Our case study is MASCOT, a pair of master--slave robotic arms used at the \gls{ccfe} in the UK to service the Joint European Torus (JET) nuclear fusion reactor. JET is operated by the \gls{ukaea} under contract from the European Commission, and exploited by the EUROfusion consortium of European Fusion Laboratories\footnote{\url{https://www.euro-fusion.org}}. The slave arms mirror the movements of the master arms, which are manually controlled by a human operator, enabling human operators to work remotely inside JET. 

Over 350 tools have been adapted to fit on the end of the MASCOT manipulators. Tools can be changed during operations, picked from a `tool box' that is also moved inside the reactor. MASCOT can be used to install, clean, and repair components inside the reactor. One major project used MASCOT for 18 months, two shifts a day, for replacement of tiles inside the reactor.

A programme is underway to update the system to `MASCOT 6', which includes adding an autonomous mode to perform some basic repetitive operations without human intervention. This update has prompted a new safety design\footnote{The safety design is confidential, so we are unable to make it publicly available. The relevant part of the document is available to reviewers in Appendix \ref{app:safetyDesign}} upon which we base our formal model. The key safety concerns relate to keeping the human operator safe during operation and maintenance.

The rest of this section is structured as follows. In \S\ref{sec:mascotSafetyDesign} we describe the pertinent parts of the MASCOT 6 Safety Design document. \S~\ref{sec:modelling} describes how the safety design was modelled. Finally, \S\ref{sec:modelVerification} describes how the model was validated and verified.

\subsection{MASCOT 6 Safety Design}
\label{sec:mascotSafetyDesign}



The MASCOT 6 safety design identifies that autonomous movement in the slave-arms would be mirrored in the master-arms, causing a hazard to the operator. Mitigating this risk is the key concern of the safety subsystem described in the safety design, which introduces two modes of operation: hands-on and autonomous, with a different speed limit for the MASCOT arms in each mode.

The safety design defines seven safety `concepts', which are the components of the safety control system. The safety properties are mixed in with the descriptions of the concepts, often as sentences that abstractly state that the system should or should not do something. 

\noindent We directly model the following six components:
\begin{compactenum}
\item Emergency/Protective Stop, which controls both manually (Emergency) and automatically (Protective) stopping the system;
\item Safe State Key Switch, which initiates an emergency stop, triggered from the work area of either the master- or slave-arms;
\item Master Commissioning Mode Key Switch, which enables the operator to put the master arms into a Commissioning State, for repairs etc.;
\item Slave Commissioning Mode Key Switch, which enables the operator to put the slave arms into a Commissioning State (different to that of concept 3);
\item Master Safe Speed Monitor, which monitors the speed of the master arms, and raises a Protective Stop if the speed limit is broken; and,
\item Master Hands-on Mode Monitor, which toggles the safety system between the Hands-on and Autonomous modes, in response to the foot pedal switch.
\end{compactenum}
\noindent The final component (which is not modelled explicitly) is an output that indicates the control system is still active, which is indirectly shown by other components in the model.

The \gls{ssm-text} and the \gls{homm-text} are the core components of the safety subsystem, cooperating to enforce the speed limit relevant to the current mode. The speed limit when the system is in autonomous mode is half that of the speed limit when in hands-on mode. 

The \gls{homm-text} monitors the foot pedal and tells the \gls{ssm-text} what mode the subsystem is in. The \gls{ssm-text} checks the speed and issues a protective stop if that mode's speed limit is broken. The other components of the safety system interact with these core components, adding extra complexity to the system. A user can trigger an emergency stop at any time, using the Safe State Key Switch. If the system is in the Master Commissioning Mode, then a protective stop is not issued if the speed limit is broken. If the Slave Commissioning Mode Key Switch is used, then the system performs an emergency stop (as with the Safe State Key Switch) and then allows power to some parts of the slave-arms.

Interactions between these six components make manual analysis of the whole system very difficult. Formalising the components makes the natural-language descriptions into an unambiguous specification, which enables automatic checking. In \S\ref{sec:modelling}, we describe our \gls{csp} model of the MASCOT safety system. 

\subsection{Modelling Approach}
\label{sec:modelling}

This section describes our \gls{csp} model of the MASCOT 6 safety subsystem, which is the result of step 1 (\S\ref{sec:step1}) of our workflow. The model is composed of communicating processes that correspond to the safety system's components (see \S\ref{sec:mascotSafetyDesign}). The model comprises $\sim$ 810 lines over eight files (including comments, but excluding two files of non-model validation code). \gls{fdr} shows that the model contains 321 states and 804 transitions (during a determinism check). 

Four of the six safety system components are each modelled by a process; the Emergency/Protective Stop and the Safe State Key Switch are represented by a single process, because their behaviour is very closely linked and combining them produced a simpler model. 
Other modelling abstractions include, abstracting the safe state key switches in the master and slave work areas into one component. Similarly, the speed of three different joints on each of the slave arms should be monitored. Our model abstracts this to one single speed measurement, though since the speed limit for each joint is the same, this process could be replicated.

Here, we briefly describe the \gls{csp} notation relevant to the examples in this section. \gls{csp} specifications are built from (optionally parametrised) processes. A process describes a sequence of events; for example $a \then b \then \Skip$ is the process where the events $a$ and $b$ happen sequentially, followed by $\Skip$ which is the terminating process. 
An event is an instantaneous communication on a \textit{channel}. Channels enable message-passing between processes, they are synchronous, non-lossy, and may have multiple end-points; but a process may perform an event (communicate an event on a channel) internally without a cooperating process. 

Channels may declare typed parameters: {$channel~c~:~int$} declares a channel $c$ with one integer parameter. Parameters communicated on $c$ may be inputs ($c?in$), outputs ($c!out$), or a given value ($c.value$); here $in$, $out$, and $value$ are all of type $int$. Inputs can be restricted ($c?p:set$) to only parameters ($p$) in a given $set$ (here, a set of type $int$). 

$P \extchoice Q$ offers the option of either $P$ or $Q$, once one process is picked the other becomes unavailable. Additionally, processes can be composed in sequence or parallel.
CSP provides two parallel operators; in $P \parallel[ chan ] Q$, $P$ and $Q$ run in parallel, and agree to communicate on channels in the set $chan$; in $P \parallel[pChan][qChan] Q$, $P$ and $Q$ run in parallel, and agree to communicate on the channels common to the $pChan$ and $qChan$ sets. 

The safety system is represented by the $MASCOT\_SAFETY\_SYSTEM$ process, which comprises a parallel composition of 
the processes for each of the components. Each process starts by synchronising on the $system\_init$ event, which ensures that all of the components start executing at the same time. 

The model contains two helper processes, which do not represent safety system components. The $MASCOT\_SYSTEM\_STATE$ process tracks the system's state: Safe, Autonomous, Hands-On, and the Master and Slave Commissioning modes. We assume that these states are mutually exclusive, though this was not clear from the safety design. The $ATOM\_CHAINS$ process enforces certain atomic chains of events that are required by the safety properties -- for example, when the foot pedal is pressed, the next event is a change of mode.

%
\begin{figure}[t]
\begin{align*} 
HMM\_& AUTONOMOUS\_MODE = \\[-0.4em]
  &enter\_safe\_state \then HMM\_SAFE\_STATE(AM) \\[-0.4em]
  &\extchoice \\[-0.4em]
  &foot\_pedal\_pressed.True \then enter\_hands\_on\_mode \then\\ &\t1 HMM\_HANDS\_ON\_MODE \\[-0.4em]
  &\extchoice  \\[-0.4em]
  &speed?\_ \then HMM\_PAUSE(AM) \\[-0.4em]
  &\extchoice  \\[-0.4em]
  &enter\_slave\_commissioning\_state \then HMM\_SAFE\_STATE(AM)  
\end{align*}
\vspace{-2em}
\caption{An extract from the \gls{homm}, showing the process controlling the \gls{homm} in autonomous mode. \label{fig:hmmAutonomous}}
\end{figure}
\begin{sloppypar}
The \gls{homm} process starts in autonomous mode, we assume that the foot pedal is not being pressed when the system is initialised. Figure~\ref{fig:hmmAutonomous} shows the definition of the \gls{homm} in autonomous mode. The $foot\_pedal\_pressed$ event toggles the process between hands on mode and autonomous mode. In Fig.~\ref{fig:hmmAutonomous}, $foot\_pedal\_pressed.True$ takes the \gls{homm} process into hands on mode ($HMM\_HANDS\_ON\_ MODE$). 
\end{sloppypar}

Two events, $enter\_safe\_state$ and $enter\_slave\_commissioning\_state$, trigger a change to the safe state, which is controlled by the $HMM\_SAFE\_STATE$ process. This can happen in either mode. In Fig.~\ref{fig:hmmAutonomous} the $HMM\_SAFE\_STATE$ process is called with the $AM$ parameter, which tells it to return to the autonomous mode when leaving the safe state. A different parameter is used when in hands on mode, to return to hands on mode.

Finally, in either mode, the $speed$ event pauses the \gls{homm} so that the mode cannot be changed before the \gls{ssm} has checked the speed. This is handled by the $HMM\_PAUSE$ process, which only offers the $protective\_stop$ or $speed\_ok$ events. In Fig.~\ref{fig:hmmAutonomous}, $HMM\_PAUSE$ is called with the $AM$ parameter to tell it to return to the autonomous mode when resuming.

\begin{figure}[t]
\begin{align*} 
SSM\_&AUTONOMOUS\_MODE = \\[-0.4em]
  &enter\_hands\_on\_mode \then SSM\_HANDS\_ON\_MODE \\[-0.4em]
  &\extchoice  \\[-0.4em]
  &speed?s:AutonomousSafeSpeeds \then speed\_ok \then SSM\_AUTONOMOUS\_MODE \\[-0.4em]
  &\extchoice  \\[-0.3em]
  &speed?s:AutonomousUnSafeSpeeds \then protective\_stop \then \\[-0.4em]
	&  \t2 enter\_safe\_state \then  SSM\_SAFE\_STATE(AM) \\[-0.4em]
  &\extchoice  \\[-0.3em]
  &enter\_safe\_state \then SSM\_SAFE\_STATE(AM) \\[-0.4em]
  &\extchoice  \\[-0.3em]
  &enter\_slave\_commissioning\_state \then SSM\_SAFE\_STATE(AM) 
\end{align*}
\vspace{-2em}
\caption{An extract from the \gls{ssm}, showing the process controlling the \gls{ssm} in autonomous mode. \label{fig:ssmAutonomous}}
\end{figure}

The \gls{ssm} process also starts in autonomous mode and toggles between that and hands-on mode, but its trigger to change mode is an $enter\_hands\_on\_mode$ or $enter\_autonomous\_mode$ event from the \gls{homm}. This allows both processes to change mode together, while \gls{homm} handles the $foot\_pedal\_pressed$ event alone. This modular design is repeated throughout, where one process handles an external event and communicates with other processes via an internal channel. 

\begin{sloppypar}
Figure~\ref{fig:ssmAutonomous} shows the definition of the \gls{ssm} in autonomous mode. The $enter\_hands\_on\_mode$ event is driven by the \gls{homm}, and triggers the \gls{ssm} to change to hands on mode. The events $enter\_safe\_state$ or $enter\_slave\_commissioning\_state$ trigger a change to the safe state; similarly to the \gls{homm} this calls a process with the $AM$ parameter so that the \gls{ssm} returns to autonomous mode when leaving the safe state. 
\end{sloppypar}

The \gls{ssm}'s role is to check the parameter of the $speed$ event. 
In Fig~\ref{fig:ssmAutonomous}, if the speed parameter is in the set of safe speeds for the autonomous mode, $speed?s:AutonomousSafeSpeeds$, then the response is $speed\_ok$; if it is in the set of unsafe speeds, $speed?s:AutonomousUnSafeSpeeds$, then the response is $protective\_stop$ and moving to the safe state. This drives the \gls{homm}, as described earlier. 

Other parts of our model make use of more complex \gls{csp} operators. The full model is available online\footnote{The model is available at \url{https://doi.org/10.5281/zenodo.3932004}}. The next section describes how we validate the model against the safety design and verify that it preserves the safety properties

\subsection{Model Validation and Verification}
\label{sec:modelVerification}

This section describes applying step 2 of our workflow (\S\ref{sec:step2}) to: validate the model against the MASCOT 6 Safety Design (\S\ref{sec:mascotSafetyDesign}), and verify that it preserves the safety properties. This step ensures that the model correctly represents the safety design, so that it can be used as the \gls{rv} oracle.


First, we validate the model to show that it represents the safety design. We use \gls{fdr} to automatically check for deadlock, divergence, and determinism, which can identify undesirable behaviour; we also use \gls{fdr}'s Probe tool to step through a process, choosing the order of events. Both are invaluable tools for specification debugging. This is an iterative process where: the model is checked, compared to the safety design, and edited (where needed). This step  provides confidence that the model accurately captures the safety concepts. 

Specification debugging revealed a problem in the safety design where the \gls{homm} could change mode before the \gls{ssm} had reacted to an unsafe speed. For example, if the system was in autonomous mode and the \gls{ssm} (Fig.\ref{fig:ssmAutonomous}) receives a $speed$ event with an unsafe speed, it should perform $protective\_stop$ and tell the \gls{homm} to $enter\_safe\_state$, but the \gls{homm} (Fig.\ref{fig:hmmAutonomous}) had the chance to perform  $foot\_pedal\_pressed.True$, before receiving $enter\_safe\_state$ This deadlocked the two processes, with \gls{homm} waiting for $enter\_hands\_on\_mode$ to be available in \gls{ssm}, and \gls{ssm} waiting for $enter\_safe\_state$ to become available in \gls{homm}. 

This problem allowed the speed limit to be changed, after a potentially unsafe speed had been recorded but before the system could enter the safe state. Discussions with the MASCOT team at \gls{ccfe} confirmed that this was incorrect behaviour. Once it was confirmed that the \gls{ssm} should take precedence over the \gls{homm}, we added a process to ensure this behaviour. Identifying bugs like this shows the utility of careful specification debugging. 

Next, we show that the model implements the safety system's requirements by verifying model against the requirements, as captured by the safety properties. In \gls{csp}, a safety property is specified as a process. The \gls{csp} model checker, \gls{fdr}, checks that the safety specification is refined by the system specification; that is, that the system implements the safety property. Again, this is an iterative process, where if checking a property failed, the the cause of the failure was investigated and the model was updated to fix this bug. In turn, this required edits that led back to the validation step.


We use the \gls{homm} process as an illustrative example of the model verification. We identified three safety properties in the safety design for the \gls{homm}. The simplest of these is: ``The monitored foot pedal is the only way for Hands-on Mode to be entered", which is captured by the safety specification:
\begin{align*}
HMM1 = foot\_pedal\_pressed.True \then enter\_hands\_on\_mode \then HMM1
\end{align*}
The $HMM1$ process allows the foot pedal to be pressed ($foot\_pedal\_pressed.True$) and then enters hands on mode ($enter\_hands\_on\_mode$). We use \gls{fdr} to check that the $MASCOT\_SAFETY\_SYSTEM$ process implements $HMM1$.

The next identified requirement is a little more complicated: \textit{``Autonomous mode is entered if the control system indicates it is no longer in Hands-on Mode''}. This is modelled by the safety specification:
\begin{align*}
HM&M2 = \\
 &foot\_pedal\_pressed.False \then enter\_autonomous\_mode \then HMM2 \\[-0.4em]
&\intchoice\\[-0.5em]
 & enter\_autonomous\_mode \then HMM2
\end{align*}
$HMM2$ allows a non-deterministic choice ($\intchoice$) between detecting that the foot pedal has not been pressed, then entering autonomous mode; or entering autonomous mode, which allows the system to perform this event when triggered by something else that doesn't affect this safety specification. We use $\intchoice$ to allow $HMM2$ to refuse one of the choices, ensuring that the model can enter autonomous mode for other reasons then the foot pedal not being pressed. Again, we use \gls{fdr} to check that the model of the system implements $HMM2$.


We check that the model implements similar safety specifications for the processes that capture the other safety concepts, which provides confidence that the model implements the requirements in the safety design. The validation step gives  crucial confidence that the safety design is modelled correctly, and gives the benefit of checking the safety document `for free'.

\section{Toolchain Evaluation}
\label{sec:eval}

This section describes the evaluation\footnote{The log files and results are available at \url{https://doi.org/10.5281/zenodo.3932004}} of the {\Varanus} toolchain (\S\ref{sec:approach}) using offline RV of the case study presented in \S\ref{sec:mascot}. This corresponds to steps 3 (\S\ref{sec:step3} and 4 (\S\ref{sec:step4}) of our workflow. All of the results are from running Python 2.7.18 and \gls{fdr} 4.2.7 on a PC using Ubuntu 20.04.02, with an Intel Core i5-3470 3.20 GHz × 4 CPU, and 8 GB of RAM.

This work focuses on offline \gls{rv}. As previously mentioned, {\Varanus} is also capable of \textit{online} \gls{rv} but when trailed, the response time was too high for effective use online. We discuss this limitation in \S\ref{sec:discussion} and intend to investigate mitigations as future work. 

We trial {\Varanus}'s response times on constructed traces: first, stress-testing {\Varanus}, with increasingly long, semi-random traces; then, checking a set of scenarios that might occur during a hypothetical mission. The scenarios were based on MASCOT log files and personal correspondence with the MASCOT team at \gls{ccfe}. This approach was taken because MASCOT 6 is still under development, so it was not possible to test \gls{rv} directly, or to compare the execution times of MASCOT with and without monitoring. 

We built a Python test harness that checks both the stress-testing and scenario traces directly in \gls{fdr}'s API and using {\Varanus} for offline \gls{rv}. The \gls{fdr} API is called by the test harness and only the checking time is logged. For offline \gls{rv}, the time taken to read the whole trace from a file and get the result from checking it in \gls{fdr} is logged. These trials give us an idea of the scalability of the model checking in \gls{fdr}, and the response times for {\Varanus} checking traces from a log file.

\begin{sloppypar}
The stress-test traces begin with $system\_init$ and add from 10 upto 100,000 $foot\_switch\_pressed$ and $speed$ events. We used a Python script to generate the traces so as to make sure that the $foot\_switch\_pressed$ channel's parameter toggled between $true$ and $false$, and the $speed$ channel's parameter was always below the autonomous (lower) speed limit. This was to ensure that the entire trace would be checked, instead of a counterexample being generated midway through. 
\end{sloppypar}
\begin{table}[t]
\centering
\begin{tabular}{l|l|l|l}
\hline
Trace Length & FDR (s) & \Varanus~Offline (s) & Difference (s)  \\ \hline \hline
11     		& 0.02	& 0.11	& 0.09 \\ \hline
101    		& 0.03	& 0.13	& 0.10	\\ \hline
1001   		& 0.17	& 0.31	& 0.14	\\ \hline
10,001  	& 1.53	& 2.02	& 0.49	\\ \hline
100,001 	& 17.97	& 20.60	& 2.63	 \\
\hline
\end{tabular}
\caption{Results, in seconds, for checking traces of length 11, 101, 1001, 10,001, and 100,001 in \gls{fdr}'s API and {\Varanus}, and the difference between these times \label{tab:stressResults}}
\end{table}
Table~\ref{tab:stressResults} shows the mean result in seconds of running 10 $[has~trace]$ checks using \gls{fdr}'s API and {\Varanus}, on different trace lengths. It also shows the difference between these times.  The checking times in both cases rises with the length of the trace. The results for traces of 1001 events or fewer, are 0.31s or less, each with an overhead of less than 0.15s. 

The scenario traces represent 13 different `attempts' at a hypothetical mission using MASCOT to replace insulating tiles on the inside of the reactor. Notes in MASCOT logs suggest that this consists of repeated actions on a group of tiles, such as: removing, installing, and tightening the bolts, etc. The data in the logs shows mainly low velocities, with some spikes that tend to appear much later on, possibly as the task becomes more difficult, this is mirrored in the scenarios.

Table~\ref{tab:scenarios} summarises the scenario traces, which exercise different features of the safety system under a variety of circumstances. They test different combinations of events, mixing changes between autonomous and hands on modes, different speeds, and other safety system components. Between them, the scenarios cover all of the safety concepts modelled from the safety design.
\begin{table}[t]
\begin{tabularx}{\textwidth}{l|l|X}
\hline
No. & Concept(s) & Description \\ \hline \hline
1	   & 1, 5, and 6  & Operator stays in hands on mode, speed stays below limit.\\ \hline
2	   & 1, 5, and 6  & Operator stays in hands on mode, speed exceeds limit and tries to continue (causes a failure). \\ \hline
2a     & 2                & Instead of the failure in Scenario 2, the system handles the broken speed limit, then resets, restarts, and finishes the mission. \\ \hline
2b     & 2                & Instead of the failure in Scenario 2, the system handles the broken speed limit, the safe state key is removed (to allow servicing). Then the key is returned, the system is reset, restarted, and operation continues. \\ \hline
3	   & 1, 5, and 6  & Operator switches to autonomous mode after collecting tools, speed stays below limit. \\ \hline
4      & 1, 5, and 6  & Operator switches to autonomous mode after collecting tools, speed exceeds limit and tries to continue (causes a failure). \\ \hline
4a     & 2              & Instead of the failure in Scenario 4, the system handles the broken speed limit, then resets, restarts, and finishes the mission. \\ \hline
4b 	   & 2              &  Instead of the failure in Scenario 4, the system handles the broken speed limit, the safe state key is removed (to allow servicing). Then the key is returned, the system is reset, restarted, and operation continues. \\ \hline
5 	   & 2                & The Safe State Key is used to trigger an emergency stop. Then the system is reset, restarted, and the mission is completed. \\ \hline
6      & 3                & System enters Master Commissioning Mode. After some unmonitored movements (not triggering protective stop), Safe State Key is used to enter Safe State, and system is reset. \\ \hline
7 	   & 4 				& System enters the Slave Commissioning Mode, where no speed events are registered. Then Slave Commissioning Mode is disabled, again using the Slave Commissioning Mode key. \\
\hline
\end{tabularx}
\caption{Table of scenarios used to evaluate the {\Varanus} toolchain, showing the Scenario Number, the identifier of Safety Concepts that it tests, and its description.\label{tab:scenarios}}
\end{table}

Table~\ref{tab:missionResults} shows the mean result in seconds of running 10 $[has~trace]$ checks in \gls{fdr}'s API, and using {\Varanus} for offline \gls{rv}}, for each scenario trace. It also shows the length of the trace and the difference between these times.  Scenarios 2 and 4 are built to fail, so while they are each 193 events long they fail after 84 events and 66 events, respectively. 
\begin{table}[t]
\centering

\begin{tabular}{l|l|l|l|l}
\hline
Scenario Name & Trace Length & FDR (s) & \Varanus~Offline (s) & Difference (s) \\ \hline \hline
Scenario 1		 & 193	  	& 0.05	& 0.15	& 0.10	 \\ \hline
Scenario 2 		 & 193 (84)	& 0.04	& 0.13	& 0.10	 \\ \hline
Scenario 2a 	 	 & 155	  	& 0.04	& 0.14	& 0.09	 \\ \hline
Scenario 2b		 & 158	  	& 0.04	& 0.14	& 0.10	 \\ \hline
Scenario 3		 & 193	  	& 0.05	& 0.15	& 0.10	 \\ \hline
Scenario 4		 & 193 (66)	& 0.03	& 0.13	& 0.10	 \\ \hline
Scenario 4a		 & 200	  	& 0.05	& 0.15	& 0.10	 \\ \hline
Scenario 4b		 & 203	  	& 0.05	& 0.16	& 0.11	 \\ \hline
Scenario 5		 & 201	  	& 0.05	& 0.16	& 0.11	 \\ \hline
Scenario 6		 & 45	  	& 0.03	& 0.13	& 0.10	 \\ \hline
Scenario 7		 & 10	  	& 0.02	& 0.12	& 0.10	 \\ 
\hline
\end{tabular}
\caption{Results, in seconds, of checking the scenarios in Table~\ref{tab:scenarios}, in \gls{fdr}'s API and {\Varanus}, and the difference. Each result is a mean over 10 checks. Scenarios 2 and 4 are built to fail, the length of the passing trace is noted in brackets. \label{tab:missionResults}}
\end{table}

%

{\Varanus} adds an overhead of $\sim$ 0.10s, and the maximum checking time is 0.16s, though this is no surprise since the traces are all less than 1001 events long. Again, the overhead varies slightly with trace length, but since the lengths of the scenario traces only has are range of 193 events, the range of the differences is only 0.02s.  Possibly these trace lengths aren't indicative of those that would be produced by MASCOT, mitigation strategies for this are discussed in \S\ref{sec:discussion}.

The response times for both \gls{fdr} and {\Varanus} rise with the length of the trace because both of them must process each event in the trace to conclude a verdict. There is little discernable difference in the response time for a trace that uses one safety concept (such as scenarios 5 or 6) and those use many safety concepts (such as scenarios 1, 2, 3, and 4). A more detailed study of \gls{fdr} is required to answer questions about the response times for more complex traces or assertions, this is left for future work.

\section{Discussion}
\label{sec:discussion}
An important benefit of our workflow is that it promotes specification reuse, which partly mitigates this bottleneck in using formal methods~\cite{rozier2016}. The \gls{csp} model is built for static verification, and then reused for \gls{rv}. Because {\Varanus} uses the model as the \gls{rv} oracle, if we are confident of the model's validity, we can be confident of the monitor's validity. 

The modelling and validation steps of our workflow may reveal problems in the original documentation, `for free'. Formalisation forces ambiguous natural-language descriptions to be clarified, and model checking examines the system's design. As previously mentioned, the model validation step (\S\ref{sec:modelVerification}) highlights an omission in the MASCOT 6 safety design that caused a spurious deadlock in the model. Our approach provides an example of the utility of formal methods even if only applied to some stages of a system's development lifecycle.

This paper focusses on using {\Varanus} for offline \gls{rv}, which is useful for detecting when the system violates its safety requirements. Formalising requirements and then using that formal model directly as an \gls{rv} oracle has the added benefit of easing the burden of traceability of the requirements into a system artefact. This should make updates and debugging simpler and make it easier to demonstrate to a regulatory authority that the system meets its requirements.


\gls{csp} generally allowed for easy translation of the behaviour described in the safety design. The MASCOT 6 safety system is composed of separate, but interacting, safety functions; this leads naturally into modelling the safety functions as parallel processes. The safety design often describes behaviour in a way that was easy to formalise in \gls{csp}, for example \textit{``Initiate a Protective Stop if the speed threshold is exceeded''} easily translates into: \\
\centerline{$speed?s : AutonomousUnSafeSpeeds \then protective\_stop $}
from Fig.\ref{fig:ssmAutonomous}
where the $speed$ event occurring with a parameter that is in the set $AutonomousUnSafeSpeeds$ (there is a similar set of unsafe speeds for the hands-on mode) triggers the $protective\_stop$ event. Earlier incarnations of this work attempted to convert our model into existing \textit{implementations} of \gls{csp}; two in Python (Python-CSP\cite{mount_csp_2009} and PyCSP\cite{bjorndalen_pycsp_2007}) and the other in Scala ($CSP_E$~\cite{yamagata_runtime_2016}). However, all three implementations differed slightly from `standard' \gls{csp}, which made systematic conversion difficult.



As mentioned in \S\ref{sec:eval}, {\Varanus} was trailed on online \gls{rv} but the response time is too high for effective online use. The mean (over 10 runs) online response time for Scenario 1 was 6.8s, compared to the 0.15s for offline \gls{rv} (Table~\ref{tab:scenarios}). This large overhead is because \gls{fdr}'s API is not designed for online \gls{rv} and requires the whole trace to be rechecked after each new \gls{sua} event. We intend to improve {\Varanus}'s online response times in future work.


{\Varanus}'s response times for all but the the largest stress-testing trace (100,001 events), are less than 0.50s.  The largest scenario trace is 203 events long, but the MASCOT logs contain many thousands of records. It is unclear how fast MASCOT 6 will produce events, but it is likely that the trace will eventually become too long. We can see two paths for optimisation, both left for future work. (1) Traces could be filtered, to only contain events for one safety concept; sampled, to yield shorter representative traces; or reset after the \gls{sua}'s behaviour cycles back to the initial state. (2) Investigate optimisations in \gls{fdr}'s API, or replace it. 


\section{Conclusion}
\label{sec:conclusion}

This paper presents the application of a workflow and novel \gls{rv} tool ({\Varanus}) to an industrial case study. We model the safety requirements found in a natural-language safety document in \gls{csp}, and use {\Varanus} for offline \gls{rv} of the system against the \gls{csp} model. The events from the case study system are converted into a \gls{csp} trace and sent to \gls{fdr} to test if the model accepts the trace.

Our approach enables a system to be modelled and monitored using the same language, without modifications between these two activities. Reusing the model like this is helpful, since modelling is often the bottleneck in using formal methods~\cite{rozier2016}. It also provides validation of the safety documents `for free'. \gls{csp} is designed for specifying processes, so it is suited to capturing imperative descriptions of behaviour. In the author's experience, modelling requirements for or by software or mechanical engineers in \gls{csp} is relatively easy.

We demonstrate {\Varanus} and the workflow, on the MASCOT teleoperation system. We model the safety system and safety properties from an English-language safety design document. We stress-test {\Varanus} to find the response times for traces of different lengths. This shows that traces under 1000 events are checkable in less than 1s. MASCOT currently does not  implement the safety system so we evaluate {\Varanus} on scenarios constructed from MASCOT logs and personal communication with members of the MASCOT team at \gls{race}.

For future work, we intend to investigate improvements to {\Varanus}'s \textit{online} response times. As previously mentioned, the \gls{fdr} API is not designed for online \gls{rv}, and our current approach adds an unacceptable time overhead.  We also intend to apply the workflow to safety \textit{cases} and examine what benefits can be gained from their structure. If possible, we want to automate the extraction of safety properties from a safety case -- or at least highlight the likely nodes in safety case that contain the safety properties to ease the modelling workload. 

\todo[inline, color=red]{This should be on page 15}
\bibliographystyle{splncs03}
\bibliography{Paper:CSP-Monitoring.bib} 
\todo[inline, color=red]{This should be on page 17}

%
%
%
%
%

\end{document}